\documentclass[12pt]{article}

\usepackage[utf8]{inputenc}
\usepackage[pdftex]{graphicx}
\usepackage[lofdepth,lotdepth,labelformat=empty]{subfig}
\graphicspath{{./figures/}{./flowchart/}}

\usepackage{colortbl}
\usepackage[margin=1in]{geometry}
\usepackage{setspace}
\usepackage{sectsty}
\allsectionsfont{\singlespacing}

\usepackage{lscape}
\usepackage{multirow}
\usepackage{ctable}
\usepackage{amsmath}
\usepackage{natbib}
\usepackage{color}
\usepackage{eurosym}
\usepackage{booktabs}
\usepackage{dcolumn}
\newcolumntype{.}{D{.}{.}{-1}}
\usepackage[toc,page]{appendix}

\newcolumntype{C}[1]{>{\centering\let\\\tabularnewline}p{#1}}
\newcolumntype{R}[1]{>{\raggedleft\let\\\tabularnewline}p{#1}}
\newcolumntype{L}[1]{>{\raggedright\let\\\tabularnewline}p{#1}}

\usepackage{color}
\definecolor{darkred}{rgb}{0.5,0,0}
\definecolor{darkgreen}{rgb}{0,0.5,0}
\definecolor{darkblue}{rgb}{0,0,0.5}

\usepackage{hyperref}
\hypersetup{colorlinks,
 linkcolor=darkblue,
 filecolor=darkgreen,
 urlcolor=darkblue,
 citecolor=darkblue,
 pdftitle={},
 pdfauthor={},
 pdfsubject={}
}

\begin{document}

\title{Transfer Topic Labeling with Domain-Specific Knowledge Base: An Analysis of UK House of Commons Speeches 1935--2014}

\author{
Alexander Herzog\\
  Clemson University\\
  \href{mailto:aherzog@clemson.edu}{aherzog@clemson.edu}
\and
Peter John\\
  King's College London\\
  \href{mailto:peter.john@kcl.ac.uk}{peter.john@kcl.ac.uk}
\and
Slava Jankin Mikhaylov \\
 University of Essex\\
  \href{mailto:s.mikhaylov@essex.ac.uk}{s.mikhaylov@essex.ac.uk}\\  
 }

\date{}


\maketitle

\begin{abstract}

Topic models are widely used in natural language processing, allowing researchers to estimate the underlying themes in a collection of documents. Most topic models use unsupervised methods and hence require the additional step of attaching meaningful labels to estimated topics. This process of manual labeling is not scalable and suffers from human bias. We present a semi-automatic transfer topic labeling method that seeks to remedy these problems. Domain-specific codebooks form the knowledge-base for automated topic labeling. We demonstrate our approach with a dynamic topic model analysis of the complete corpus of UK House of Commons speeches 1935-2014, using the coding instructions of the Comparative Agendas Project to label topics. We show that our method works well for a majority of the topics we estimate; but we also find that institution-specific topics, in particular on subnational governance, require manual input. We validate our results using human expert coding.


\end{abstract}
\thispagestyle{empty}

\newpage

\doublespacing

\section{Introduction}\label{sec:introduction}

Political science scholars working with large quantities of textual data are often interested in discovering latent semantic structures in their document collections. Examples include legislative debates, policies, media content, manifestos, and open-ended survey questions. Manual coding of such data is extremely time-consuming and expensive, and does not scale well with the expanding size of the corpora. In practice, document summarization is routinely carried out using variations of probabilistic topic models \citep{BleiNgJordan2003}. However, semantic understanding of resulting topics still requires human involvement and thus a set of discretionary decisions that need validation and ensure transparency. 

A key feature of political corpora is the evolution of semantic structures over time, which is not fully accounted for in existing methods. Politicians and other decision-makers choose to pay attention to functional areas of specialization, such as health or education policies, which reflect their interests, career experience and goals, and demands for representation. The agenda of the legislature and the content of debates shifts across these topics over time, according to functional pressures and agenda-setting in the media and from public opinion. This is known as concept drift \citep{gama2014survey}. In natural language processing (NLP), dynamic topic models are often used to capture the evolution of content structure over time \citep{DTM}. Concept drift in political documents comes from changing content of the texts and presentation style (e.g. compare UN General Debate speeches by Obama and Trump), as well as from the adaptive behavior of politicians \citep{baturo2017understanding}. The vocabulary used to express topics may change over time. Human coders are intuitively better placed to pick up the change in meaning in political texts, while machine coding is often faulted with failures to detect semantic change \citep{albaugh_comparing_2014}.  

We present a method that automatically transfers existing domain-specific knowledge base for topic labeling. We show that our method works well under the concept drift in document summarization. We illustrate the performance of our method applying dynamic topic modeling of the debates in the UK House of Commons from 1935 to 2014, and labeling the topics using the coding manual of the Comparative Agendas Project (CAP) \citep{bevan2014}. We validate our results using human labeling of the topics by the CAP expert coders. Our method applies more generally and can be easily extended to other areas with existing domain-specific knowledge base, such as party manifestos, open-ended survey questions, social media analysis, and legal cases. Using our method, researchers in these fields can be more confident that the building blocks of their models are not an artefact of human coding decisions from within the research process itself.

\section{Related Work}

In the absence of roll-call data that can be used for ideal point estimation, scholars have turned to legislative speech to estimate policy positions, either by focusing on selected debates \citep[e.g.][]{LaverBenoit2002,HerzogBenoit2015} or through the analysis of all speeches during a legislative term \citep{LauderdaleHerzog2016}. A parallel stream of the literature has used topic modeling to estimate the extent to which legislators speak on different topics \citep{quinn_how_2010}. Topic modeling is a class of models that estimate the underlying themes in a collection of documents. Originally proposed by \cite{BleiNgJordan2003} in their seminal paper on latent Dirichlet allocation (LDA), various extensions of LDA have been developed in the computing sciences \citep[e.g.][]{CTM,HDP,Roberts2016}. There have been several recent applications in political science \citep{Grimmer2009,Roberts2014,mueller2018reading}.  

One of the LDA extensions is the dynamic topic model (DTM) \citep{DTM}, which relaxes the assumption of LDA that documents are unordered. Instead, DTM assumes that documents are grouped into discrete time intervals (e.g., years) that exhibit different mixtures of topics, which allows topics to change over time -- both in terms of their prevalence in the corpus and in their word compositions. DTM has not seen extensive practical use with large volumes of data potentially due to scalability of the inference algorithm \citep{gropp2016scalable}. In political science there have been very few applications \citep[e.g.][]{gurciullo2015complex,greene_exploring_2017}. 

Collections of political documents spanning long periods of time exhibit a problem known as concept drift \citep{gama2014survey}. Under `not strictly stationary' data generating processes, the underlying concept that is the target variable of our prediction model may be changing over time thus affecting the predictive decision \citep[for a formal definition,  see][]{webb2016characterizing}. In political science, \cite{lowe2011scaling} develop a measurement model to address the changing nature of left-right ideological dimension.  \cite{benoit2016crowd} present a general approach to data generation using crowdsourcing that can quickly react to change of concepts like the appearance of immigration as a new dimension of party competition. A standard approach to deal with concept drift in political science NLP applications has been to analyze the data from each time interval separately. For example, recent work using annual speech data estimates separate models for each year \citep[e.g.][]{HerzogBenoit2015,baturo2013life}. DTM is a drift-aware adaptive learning algorithm that adapts to the evolution of topics over time. 

Topic labeling is a key post-processing step of all probabilistic topic models. As a general rule labels should be relevant, understandable, with high coverage inside topic, and discriminative across topics. Early research focused on generating labels by hand by using a set of top \emph{n} words in a topic distribution (so called \emph{cardinality}) learned by a topic model \citep[e.g.][]{griffiths2004finding}. An alternative approach is to implement a supervised topic modeling approach that limits the topics to a predefined set with their word distributions provided \emph{a priori} \citep{mcauliffe2008supervised,ramage2009labeled}. The former approach is not scalable, carries a high cognitive load in forming the topic concept and its interpretation \citep{aletras2017labeling}, and also suffers from a potential bias of the human labeler \citep{lau2016sensitivity}, while the latter is unable to pick up topics unknown beforehand \citep{wood2017source}.

Several automatic labeling approaches have been proposed in the literature that utilize external, contextual information. \citet{mei2007automatic} minimize the semantic distance between the topic model and the candidate label based on the phrases inside documents. \citet{lau2011automatic} utilize various ranking mechanisms of the top-\emph{n} words and candidate labels from Wikipedia articles containing these terms. \cite{bhatia2016automatic} use word embeddings to map topics and candidate labels derived from Wikipedia article titles, and then select topic labels based on cosine similarity and relative ranking measures. Word embeddings pre-trained on a large corpus like Wikipedia and deployed for topic labeling of PubMed abstracts as in \cite{bhatia2016automatic} is a simple form of general domain knowledge transfer. More generally, a machine learning framework captures the ability to transfer knowledge to new conditions, which is known as transfer learning \citep[for a survey and formal definition see][]{pan2010survey}.

\section{Unsupervised Topic Modeling with Transfer Topic Labeling}

\begin{figure}[tp]
\centering
\includegraphics[width=\textwidth]{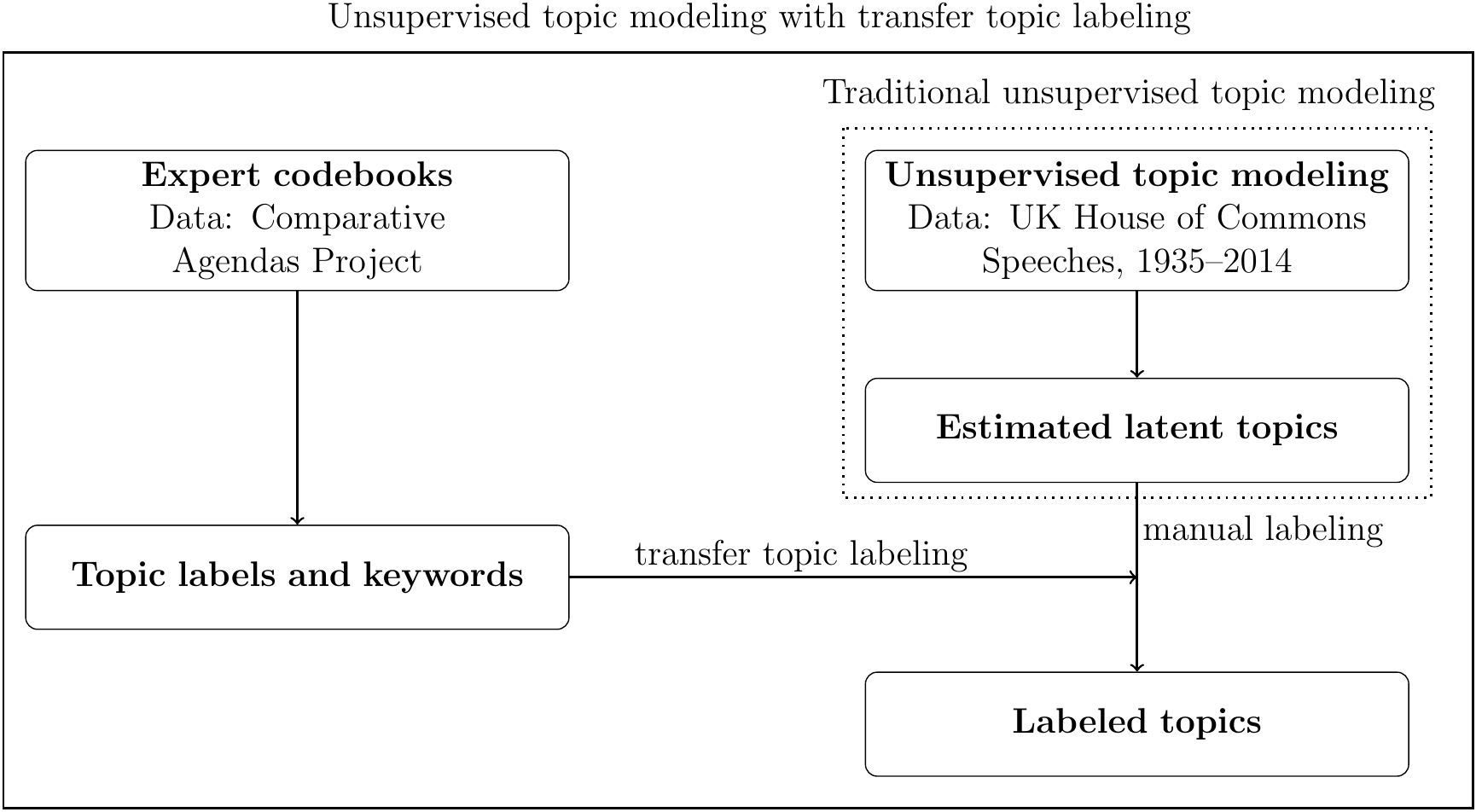}
\caption{Illustration of the application of transfer learning for semi-automated labeling of estimated topics.}
\label{fig:flowchart}
\end{figure}

Our main idea is illustrated in Figure~\ref{fig:flowchart}. The dotted box on the right-hand side illustrates traditional unsupervised topic modeling, which stops with estimated latent topics that need manual labeling. In our approach, we use outside expert codebooks to extract topic labels and associated keywords, which we then use to automatically label the estimated latent topics. Retaining human-in-the-loop allows for adjustment of the labels for specific domains with sparse coverage in the source knowledge base. Hence, we use the term semi-automated topic label in this paper.

In the remainder of this section, we demonstrate the utility of this approach with speeches from the UK House of Commons over the time period from 1935 to 2014. We first explain how we have estimated latent, dynamic topics from the speeches. We then discuss how we have used the codebooks from the Comparative Agendas Project (CAP) \citep{bevan2014} as existing knowledge base to transfer topic labels.

\subsection{Estimating Dynamic Topics from House of Commons Speeches, 1935--2014}

Our data consist of the complete record of debates from the UK House of Commons during the time period  1935--2014. All debates and information about speakers were downloaded from TheyWorkForYou,\footnote{\url{https://www.theyworkforyou.com/}} a transparency website that provides access to parliamentary records and information about MPs. All data were downloaded in XML format and further processed in Python.

The full data set consists of about 4.3 million floor contributions with an average of 49,720 contributions per year (min=17,280, max=118,500, sd=17,596) and a total of 117,914 unique words. Within each session, we combined each MP's contributions into a single text, excluding contributions that concern the rules of procedure or the business of the House, such as the reading of the parliamentary agenda or formal announcements. We also removed the traditional prayer at the beginning of each sitting and all contributions and announcements by the Speaker. 

As part of the pre-processing, we applied stemming, removed words that appeared less than 50 times and in fewer than 5 documents, removed punctuation, numbers, symbols, stopwords, hyphens, single letters, and a custom list of high-frequency terms.\footnote{We used the fairly extensive MySQL stopword list, which includes more than 500 words. Our custom list includes the following words: hon, mr, member, members, bill, minister, prime, government(s), friend, year(s), gentleman, gentlemen.} The final data set from which we estimate topics includes 47,524 documents (i.e., an MP's concatenated speeches during a session) and 19,185 unique words.

We used the dynamic topic model (DTM) by \cite{DTM} to estimate topics from the speech data. Like any unsupervised topic model, DTM requires setting the number of topics \emph{a priori}. We followed the standard in the literature and picked the number of topics based on semantic coherence and exclusivity \citep[c.f.,][]{Roberts2016}. Based on these two metrics, we selected the model with 22 topics.\footnote{Further details and references regarding model selection are provided in the supplementary materials, Section A.}

\subsection{Extracting Topic Labels and Keywords from Expert Codebooks}

We used coding instructions from the Comparative Agendas Project (CAP) \citep{bevan2014} as external source to extract topic labels and associated keywords. We selected the CAP because we expected the majority of parliamentary speeches to be on topics related to public policy-making. This attention to policy topics is the central interest of the policy agendas set of projects, which have been active since the late 1990s \citep[e.g.][]{baumgartner_comparative_2013}. What has been called the policy agendas code frame is a fuller articulation of the policy topics ideas with a larger number of major topic codes, which aims at comprehensive coverage of any topic that is likely to appear.

While our demonstration of transfer topic labeling is limited to the CAP codebooks, we note that our method can be easily extended to other codebooks as long as they include written coding instructions or vignettes. Further, as we will demonstrate below, our method for matching topic labels to estimated latent topics produces goodness-of-fit measures for each match, which allows to evaluate how well the topics derived from a codebook capture the estimated latent dimensions. 

The codebook for the UK Policy Agendas Project includes 19 major topics with subtopics.\footnote{This codeframe for the UK is summarized in \cite{john_policy_2013} and on the UK project website (http://www.policyagendas.org.uk/).} For each subtopic, the CAP codebook provides written examples of what is being included in each category. For example, category ``1. Macroeconomics -- 100: General domestic macroeconomic issues'' is described as follows:

\begin{quote}
\emph{the government's economic plans, economic conditions and issues, economic growth and outlook, state of the economy, long-term economic needs, recessions, general economic policy, promote economic recovery and full employment, demographic changes, population trends, recession effects on regional and local economies, distribution of income, assuring an opportunity for employment to every person seeking work, standard of living.}
\end{quote}

Because the descriptions of CAP subtopics are relatively short, we combine all subtopics under a major topic label into a single document. We then apply  \emph{tf-idf} weighting to generate 19 weighted word lists (one for each major topic label), where the weight on each word reflects its importance to a topic label.\footnote{Before calculating \emph{tf-idf} weights, we applied the same pre-processing rules that we applied to the speech data to increase the similarity between the two vocabularies.}  Table~\ref{tab:policy_agenda_topics} provides an overview of the 19 topics together with their ten highest ranked words.

\begin{table}[tp]
\centering
\caption{Overview of CAP Topics} \label{tab:policy_agenda_topics}
\begin{footnotesize}
\begin{tabular}{p{0.25\textwidth}p{0.7\textwidth}}
\toprule
\textbf{Policy agenda topic} & \textbf{Top ten words based on \emph{tf-idf} weighting} \\
\midrule
Macroeconomic Issues & tax, inflat, index, treasuri, fiscal, price, taxat, unemploy, bank, gold \\
Civil Rights & discrimin, asylum, immigr, equal, right, citizenship, minor, age, refuge, freedom \\
Health & healthcar, care, health, medic, drug, coverag, nurs, provid, alcohol, mental \\
Agriculture & agricultur, farm, anim, food, livestock, produc, crop, erad, fisheri, diseas \\
Labor and Employment & employ, labour, job, migrant, youth, worker, employe, workplac, work, train \\
Education and Culture & educ, student, school, art, vocat, higher, secondari, teacher, grant, learn \\
Environment & water, pollut, environment, wast, hazard, conserv, emiss, climat, municip, air \\
Energy & electr, gas, energi, coal, oil, power, natur, nuclear, fuel, gasolin \\
Transportation & highway, transport, rail, truck, bus, road, ship, aviat, speed, air \\
Law and Crime & crime, crimin, drug, justic, traffick, polic, juvenil, sentenc, court, offend \\
Social Welfare & benefit, elder, volunt, social, food, welfar, incom, contributori, meal, lunch \\
Community Development, Planning and Housing & hous, mortgag, urban, tenant, veteran, low, homeless, citi, rural, tenanc \\
Banking and Finance & small, bankruptci, copyright, busi, patent, consum, mortgag, tourism, sport, mutual \\
Defense & defenc, weapon, arm, intellig, militari, forc, reserv, veteran, armi, war \\
Space Science & scienc, space, radio, communic, satellit, tv, launch, telecommun, broadcast, research \\
Foreign Trade & trade, export, tariff, import, invest, exchang, duti, competit, u.k, restrict \\
International Affairs and Foreign Aid & european, soviet, east, u.n, africa, u.k, peac, polit, europ, treati \\
Government Operations & postal, legislatur, execut, minist, employe, elect, census, elector, offici, prime \\
Public Lands, Water Management & indigen, land, park, convey, histor, water, forest, monument, memori, reclam \\
\bottomrule
\end{tabular}
\end{footnotesize}
\end{table}

\subsection{Transfer Topic Labeling}

We transfer topic labels from the CAP to the estimated latent topics through a pair-wise matching procedure that finds the most similar CAP topic word list for each latent dimension. For the CAP topics, the word lists are the \emph{tf-idf}-weighted word lists discussed above. For the dynamic topic model, we construct one word list for each of the 22 estimated latent topics.\footnote{Additional information on our implementation of transfer topic labeling is provided in in the supplementary materials, Section B.}

To identify the best matching topics, we use the Jaccard index, which is a widely used set-based similarity measure.\footnote{We also used an alternative approach using the ROUGE F1 metric frequently used in document summarization and machine translation literature. The results using Jaccard and ROUGE were identical.} For two sets $A$ and $B$, the Jaccard index is defined as

\begin{equation}\label{eq:jaccard}
J(A,B) = \frac{|A\cap B|}{|A\cup B|}
\end{equation}

where the numerator is the size of the intersection between $A$ and $B$, and the denominator is the size of the union of the two sets. The Jaccard index is bound between 0 and 1, with higher numbers indicating greater overlap between two sets. We calculate the Jaccard index for each pair of word lists consisting of one CAP topic and one estimated DTM topic. Using the highest Jaccard value results in 19 unique matches where the CAP label is transferred to the estimated topic. 

As a validation exercise we recruited a group of CAP experts to label the word lists for each topic according to the CAP categorization. Seventeen experts who participated in this exercise could submit two choices of the labels (most appropriate and second most appropriate). We assess the quality of expert labeling using Fleiss' kappa measure of inter-coder agreement. We also calculate proportion of experts who agree with the automatically selected topic label as their first or second choice. We provide additional information on our expert coding validation exercise in the supplementary materials, Section C.

Table~\ref{tab:labeled_topics} provides an overview of the 22 estimated dynamic topics together with their Jaccard index, the matched CAP topic label, topic label selected by experts, proportion of experts who selected the same topic label as the transfer-learning approach with the first or second choice, Fleiss' kappa inter-coder agreement, and the top 20 words from each DTM topic. Three CAP categories from Table~\ref{tab:policy_agenda_topics} have not been matched: Environment, Space Science, and Public Lands and Water Management. It should also be noted that the lists of words conform to common understandings of what the categories mean, for example agriculture has words such as milk and meat in the list.

\begin{landscape}
\begin{table}[tbp]
\begin{tiny}
\begin{center}
\caption{DTM topics with Matched Policy Agenda Topic Labels and Comparison to Expert Coding}\label{tab:labeled_topics}
\begin{tabular}{cp{0.25\textwidth}p{0.25\textwidth}ccccp{0.58\textwidth}}
\toprule
\textbf{\#} &
\textbf{Topic Label Selected by} &
\textbf{Topic Label Selected by} &
\multicolumn{2}{c}{\textbf{Prop. Experts}} & 
\textbf{Jaccard} &
\textbf{Fleiss'} &
\textbf{Top 20 Words from Estimated Dynamic Topics} \\\cline{4-5}
\textbf{} &
\textbf{Transfer-Learning Approach} &
\textbf{Experts} &
\textbf{1st} &
\textbf{2nd} &
\textbf{Index} & 
\textbf{Kappa} &  \\
\midrule
1 & Agriculture & Agriculture & 1.00 & 0 & 0.62 & 0.81 & price, agricultur, food, suppli, ask, ration, milk, water, farmer, ministri, market, industri, fisheri, consum, sugar, beef, meat, fish, rural, increas \\[0.2em] 
2 &  Labour and Employment & Labour and Employment & 0.94 & 0 & 0.47 & 0.54 & employ, industri, polic, men, labour, worker, union, work, unemploy, area, women, trade, law, wage, crime, home, court, factori, train, case \\[0.2em] 
3 &  International Affairs and Foreign Aid & International Affairs and Foreign Aid & 0.88 & 0.06 & 0.23 & 0.82 & hous, european, question, matter, eu, committe, order, union, communiti, discuss, statement, europ, made, treati, constitut, countri, debat, point, answer, make \\[0.2em] 
4 &  Defense & Defense & 0.88 & 0 & 0.42 & 0.61 & air, defenc, forc, ministri, civil, aviat, ireland, aircraft, aerodrom, servic, northern, broadcast, imperi, airway, afghanistan, televis, iraq, corpor, offic, fli \\[0.2em] 
5 &  Community Development, Planning and Housing Issues & Community Development, Planning and Housing Issues & 0.81 & 0.06 & 0.52 & 0.68 & local, hous, author, council, build, road, work, rent, charg, plan, home, region, area, counti, rate, communiti, land, london, peopl, develop \\[0.2em] 
6 &  Government Operations & Government Operations & 0.75 & 0.12 & 0.27 & 0.40 & scotland, scottish, state, vote, elect, elector, secretari, hous, parliament, commiss, regist, parti, assembl, system, ask, gallant, peopl, glasgow, awar, devolut \\[0.2em] 
7 &  Foreign Trade & Foreign Trade & 0.69 & 0.06 & 0.32 & 0.60 & trade, hous, question, committe, industri, board, matter, export, countri, import, duti, answer, discuss, refer, presid, agreement, made, film, hope, british \\[0.2em] 
8 &  Health & Health & 0.56 & 0.44 & 0.52 & 0.52 & school, educ, health, servic, care, author, hospit, evacu, nhs, children, patient, local, board, adopt, medic, peopl, teacher, area, univers, doctor \\[0.2em] 
9 &  Transportation & Transportation & 0.56 & 0.25 & 0.32 & 0.46 & busi, london, steel, product, industri, ship, war, suppli, constitu, british, ministri, transport, research, peopl, aircraft, work, rail, vessel, firm, factori \\[0.2em] 
10 &  Energy & Energy & 0.56 & 0.19 & 0.52 & 0.43 & agricultur, coal, industri, energi, land, farmer, board, oil, farm, miner, gas, subsidi, power, water, scheme, climat, british, committe, electr, carbon \\[0.2em] 
11 & Labour and Employment & Labour and Employment & 0.50 & 0.12 & 0.13 & 0.54 & question, pension, peopl, sir, figur, work, benefit, answer, increas, million, inform, rate, refer, repli, report, cent, part, gallant, committe, matter \\[0.2em] 
12 & Energy & Energy / Labour and Employment$^1$ & 0.38 & 0.19 & 0.37 & 0.48 & coal, industri, employ, unemploy, board, area, mine, job, peopl, fuel, develop, train, electr, miner, transport, region, men, work, east, north \\[0.2em] \hline
13 & Government Operations & Law, Crime, and Family Issues & 0.31 & 0.25 & 0.18 & 0.51 & peopl, home, ask, point, speaker, hous, constitu, case, offic, polic, order, general, agre, debat, secretari, prison, man, awar, post, public \\[0.2em] 
14 & Social Welfare & Macroeconomics & 0.25 & 0.06 & 0.42 & 0.18 & secretari, state, tax, chancellor, peopl, benefit, pension, exchequ, cut, war, incom, social, hous, purchas, problem, profit, minist, compani, duti, govern \\[0.2em] 
15 & Banking, Finance, and Domestic Commerce & Social Welfare & 0.06 & 0.06 & 0.23 & 0.30 & pension, nation, price, unemploy, industri, case, assist, insur, increas, busi, benefit, compani, british, peopl, age, offic, man, widow, board, allow \\[0.2em] 
16 & Macroeconomics & Transportation & 0 & 0.25 & 0.32 & 0.46 & secretari, transport, state, railway, tax, road, industri, compani, price, bank, subsidi, commiss, vehicl, trade, nationalis, control, servic, chancellor, union, privat \\[0.2em] \hline
17 & Foreign Trade & International Affairs and Foreign Aid & 0 & 0 & 0.37 & 0.82 & countri, state, commonwealth, coloni, leagu, british, intern, unit, india, foreign, secretari, majesti, ask, syria, develop, german, south, peopl, rhodesia, world \\[0.2em] 
18 & Social Welfare & Government Operations & 0 & 0 & 0.18 & 0.40 & amend, claus, point, committe, hous, learn, move, order, debat, case, matter, word, beg, line, act, deal, provis, make, legisl, law \\[0.2em] 
19 & Social Welfare & Education & 0 & 0 & 0.27 & 0.53 & ask, secretari, state, school, educ, awar, offic, statement, war, armi, servic, make, teacher, children, men, admiralti, view, step, forc, releas \\[0.2em] 
21 & International Affairs and Foreign Aid & Transportation & 0 & 0 & 0.18 & 0.46 & ask, wale, welsh, assembl, secretari, road, transport, war, awar, state, view, north, east, railway, learn, author, step, region, number, local \\[0.2em] 
22 & Social Welfare & Government Operations & 0 & 0 & 0.27 & 0.40 & matter, question, case, sir, answer, sport, made, act, local, author, fund, inform, report, point, time, nation, person, concern, regul, servic \\[0.2em] 
23 & Civil Rights, Minority Issues, and Civil Liberties & Government Operations & 0 & 0 & 0.23 & 0.40 & ireland, northern, countri, point, peopl, polic, war, hous, speech, great, time, parti, irish, speaker, debat, issu, order, opposit, state, agreement \\[0.2em] 
\bottomrule
\end{tabular}
\end{center}
\vspace{-\baselineskip}
\emph{Note}: Column ``Prop. Experts'' is the proportion of experts who selected the same topic as the transfer-learning approach with their first or second choice. \\
$^1$ Experts were tied between topic label ``Energy'' and ``Labour and Employment''. The Fleiss' Kappa reported in this row is the average of the kappas for each label.
\end{tiny}
\end{table}
\end{landscape}

A majority of experts agreed with the automatic approach on 12 topic labels. The clearest example being the topic of agriculture (\#1) where transfer labeling and all the experts identified farming and agriculture related terms. Further four topics show sufficiently large agreement between experts across two choices and automatic labeling (\#13 government operations and \#14 social welfare). Banking topic is labeled in total by 12\% of experts, but it also shows significant disagreement across experts with Fleiss' kappa at 0.3. For macroeconomics (\# 16) a majority of experts labeled it as transport, while 25\% of the experts agreed with the automatic labeling of this topic as macroeconomics as their second choice (kappa =  0.46).  

The remaining six entries in the table show complete disagreement between our automatic approach and experts. Not a single expert assigned the same label as the transfer-learning approach. These cases are difficult to explain as they both contain varied values for Jaccard and Fleiss' kappa. The topics morph into concerns about representation and territorial identity, not in ways that are solely about these political topics, but are still connected to the policy issues that MPs talk about.  With transportation the match is for regional policies in Wales and England which is an amalgam of keywords on transport. Another crossover is for social welfare that combines with legislative procedures, which reflects the extent to which MPs focus on social welfare in asking parliamentary questions. The importance of this topic is that it comes up four times with different word formations. The experts are possibly using the government operations label for catch-all procedural issues (e.g. in \#18 and \#22), or fitting a label to a topic that is not represented in CAP like Northern Ireland (\# 23). In the latter case, the algorithm arguably more correctly applies the label of civil rights and minority issues. We provide additional validation results in the supplementary materials.

\section{Conclusion}

Treating text as data is an approach of increasing importance in political science. NLP techniques developed in the computing sciences are routinely added to methodological toolkits. Topic modeling is a favorite tool of document summarization. Political scientists often have to be creative in interpreting and labeling estimated topics; yet such labeling is also often difficult to replicate -- a sine qua non of modern political science. 

To address the deficiency of current labeling and to have a better way of accommodating changes over time, we present a new method for topic labeling. Our approach provides an automatic labeling method that transfers the wealth of substantive knowledge accumulated in political science into labeling topic models.  Our transfer labeling approach is also fully transparent and replicable that allows to bring in human expertise to bear on difficult cases. Hence we call it a semi-automatic transfer labeling approach with a human-in-the-loop.

The method can be extended to party manifestos, open-ended survey questions, social media data, and legal documents, in fact all research domains where topic models have made advances in recent years.

\clearpage \singlespacing
\bibliographystyle{apsr}
\bibliography{main}

\end{document}